\documentclass[12pt,letterpaper]{article}

\usepackage{renstyles}

\usepackage{cite}
\hypersetup{
    citecolor=green!50!black,
    linkcolor=blue!80!black
}

\newcommand{\sep}{\,||\,}
 
\title{Instance-Wise Adaptive Sampling for Dataset Construction in Approximating Inverse Problem Solutions} 
\author{
    Jiequn Han \thanks{Center for Computational Mathematics, Flatiron Institute, New York, NY 10010; jhan@flatironinstitute.org}
    \and
    Kui Ren \thanks{Department of Applied Physics and Applied Mathematics, Columbia University, New York, NY 10027; kr2002@columbia.edu}
    \and
    Nathan Soedjak \thanks{Department of Applied Physics and Applied Mathematics, Columbia University, New York, NY 10027; ns3572@columbia.edu}
}
\begin{document}

\maketitle

\begin{abstract}
We propose an instance-wise adaptive sampling framework for constructing compact and informative training datasets for supervised learning of inverse problem solutions. Typical learning-based approaches aim to learn a general-purpose inverse map from datasets drawn from a prior distribution, with the training process independent of the specific test instance. When the prior has a high intrinsic dimension or when high accuracy of the learned solution is required, a large number of training samples may be needed, resulting in substantial data collection costs. In contrast, our method dynamically allocates sampling effort based on the specific test instance, enabling significant gains in sample efficiency. By iteratively refining the training dataset conditioned on the latest prediction, the proposed strategy tailors the dataset to the geometry of the inverse map around each test instance. We demonstrate the effectiveness of our approach in the inverse scattering problem under two types of structured priors. Our results show that the advantage of the adaptive method becomes more pronounced in settings with more complex priors or higher accuracy requirements. While our experiments focus on a particular inverse problem, the adaptive sampling strategy is broadly applicable and readily extends to other inverse problems, offering a scalable and practical alternative to conventional fixed-dataset training regimes.
\end{abstract}


\maketitle
\section{Introduction}
Inverse problems represent a fundamental class of challenges across numerous scientific and engineering domains, where the goal is to infer underlying parameters or structures from observable measurements. These problems are often notoriously difficult due to their ill-posed nature, often requiring sophisticated mathematical techniques and substantial computational resources to solve effectively. Traditional optimization-based methods for challenging inverse problems are often highly sensitive to initialization and can converge to undesirable local minimizers when the initial guess is insufficiently accurate~\cite{van2013mitigating, burstedde2009algorithmic, zhou2023neural, melia2025multi}. In recent years, deep learning approaches have emerged as powerful tools for approximating solutions to inverse problems, offering the potential for significantly faster inference while achieving reasonable accuracy compared to traditional optimization-based methods; see, e.g., the recent reviews~\cite{arridge2019solving,ongie2020deep,ying2022solving} and references therein.

However, a critical limitation of deep learning approaches for inverse problems is their considerable data hunger. Training effective inverse maps from measurements to underlying parameters based on neural networks typically requires large datasets of input-output pairs~\cite{zhou2023neural,klug2023scaling,adcock2024learning}, which can be prohibitively expensive to collect and use, for instance, when each forward simulation involves solving complex partial differential equations (PDEs). As prior knowledge about the inferred parameters becomes less constrained, the data requirements become increasingly demanding because more data is necessary to sufficiently cover the parameter space. This creates a substantial obstacle for applying deep learning to realistic inverse problems with complex, high-dimensional parameter spaces.

In this paper, we introduce a novel instance-wise adaptive sampling strategy that substantially reduces the sample complexity required to train neural networks for inverse problems. Rather than learning a globally accurate inverse model over the entire parameter space, our method focuses on accurately approximating the inverse map in the vicinity of each test instance. Starting from a modestly sized base dataset used to train an initial base model, we iteratively generate additional training samples near the given test instance. This targeted data augmentation creates locally enhanced training sets that are particularly relevant to each case, enabling strong reconstruction accuracy without the computational burden of generating massive general-purpose training datasets upfront.

We demonstrate our method on an inverse scattering problem for the Helmholtz equation~\cite{colton1998inverse,kirsch2011introduction}, a challenging inverse problem with applications in radar, sonar, medical imaging, and seismic exploration. In this context, the goal is to determine the properties of an unknown heterogeneous medium by probing it with incident waves and measuring the resulting scattered waves at distant locations. Numerical experiments show that models trained with our adaptive sampling approach can achieve performance comparable to or better than models trained on datasets many times larger. For a single challenging instance, the required sample size can be reduced by one to two orders of magnitude, depending on the complexity of the parameters to infer.

The proposed instance-wise adaptive sampling strategy can also be viewed as a form of inference-time scaling in inverse problems, where computational resources for data generation are allocated more efficiently by focusing on the most relevant regions of the parameter space during inference time. A similar shift is emerging in large language models (LLMs), where further scaling of pre-training is increasingly constrained by the scarcity of data and computational resources~\cite{villalobos2024position,muennighoff2025scaling}. As a result, there is growing interest in methods that adapt model behavior or resource allocation at inference time, on a per-query basis~\cite{snell2024scaling, liu2025can, openai2024o1}.

Our perspective aligns with this philosophy and suggests a parallel path for inverse problems: dynamically tailoring data acquisition to each instance can yield high-quality solutions with far fewer samples. This can help bridge the gap between traditional optimization-based methods and purely data-driven approaches. By emphasizing the quality and relevance of training data rather than solely its quantity, our approach presents a promising direction for overcoming the data efficiency challenges currently limiting the application of deep learning to complex inverse problems.

\section{Methodology}
\label{SEC:Methodology}
We consider the general formulation of an inverse problem, where a forward operator $\mathcal{F}$ maps a parameter $q$ to a measurement $m = \mathcal{F}(q)$. Given an observed measurement $\widehat{m}$, the goal is to recover a corresponding parameter $\widehat{q}$ by solving the optimization problem
\begin{equation}
\label{eq:general_loss}
\widehat{q} = \arg\min_{q} \, \mathcal{L}(\mathcal{F}(q), \widehat{m}),
\end{equation}
where $\mathcal{L}$ is a suitable loss function measuring the discrepancy between the predicted measurement $\mathcal{F}(q)$ and observed measurement $\widehat{m}$. 

Although in many setups the parameter $q$ lives in a high-dimensional ambient space, e.g. $\mathbb{R}^{N_2}$ for some large $N_2$, it is often the case that we have \emph{prior knowledge} that $q$ lies on or close to some potentially low-dimensional manifold $\mathcal{M}$ in $\mathbb{R}^{N_2}$. In particular, the intrinsic dimension $N_1$ of the parameter manifold $\mathcal{M}$ may be much smaller than the dimension $N_2$ of the ambient space for some applications. Such prior knowledge could come either from the underlying physics or from the fact that the inverse problem is so ill-conditioned that only limited information about $q$ can be reliably reconstructed~\cite{BaRe-IP09}. In this paper, we consider two representative classes of priors: smoothness-based and geometry-based, both of which are described in detail in Section~\ref{sec:prior}.

Assuming the inverse map $\mathcal{F}^{-1}$ exists and can be well approximated and efficiently evaluated, applying it directly via $\mathcal{F}^{-1}(\widehat{m})$ provides a fast approximate solution to \eqref{eq:general_loss}. A standard data-driven approach to learning the inverse map $\mathcal{F}^{-1}$ involves first randomly sampling many parameters $q_1,\dots, q_N$ from the parameter manifold $\mathcal{M}$, and then collecting the corresponding measurements $m_1,\dots,m_N$ by applying the forward operator $\mathcal{F}$ through either simulations or experiments. One can then train a machine learning model on the dataset $\{(m_1,q_1),\dots,(m_N,q_N)\}$ to obtain an approximation for the inverse operator $\mathcal{F}^{-1}$. This learning process, however, is extremely challenging. First, because this approach is purely data-driven, the size of the dataset may need to be prohibitively large~ \cite{zhou2023neural,klug2023scaling,adcock2024learning}. In the particular context of inverse scattering problems examined in \cite{zhou2023neural}, numerical results suggest that the number of training samples required for the inverse model to achieve a certain target accuracy appears to scale \textit{exponentially} with the intrinsic dimensionality $N_1$ of the manifold $\mathcal{M}$. Second, even when large training datasets are available, the resulting optimization problem is unrealistically expensive to solve for practically relevant inverse problems~\cite{DiReZh-HNA25}.

To address the sample complexity limitations of purely data-driven approaches, we propose an instance-wise adaptive sampling strategy that progressively improves reconstruction accuracy by focusing data collection in regions of the parameter space that are most relevant to the test instance. Rather than training a single global inverse model, our method adaptively refines the model for each test measurement by sampling locally on the parameter manifold and fine-tuning on this adaptive dataset. The procedure consists of the following steps:

\begin{figure}[!htb]
\centering
\includegraphics[width=0.99\linewidth]{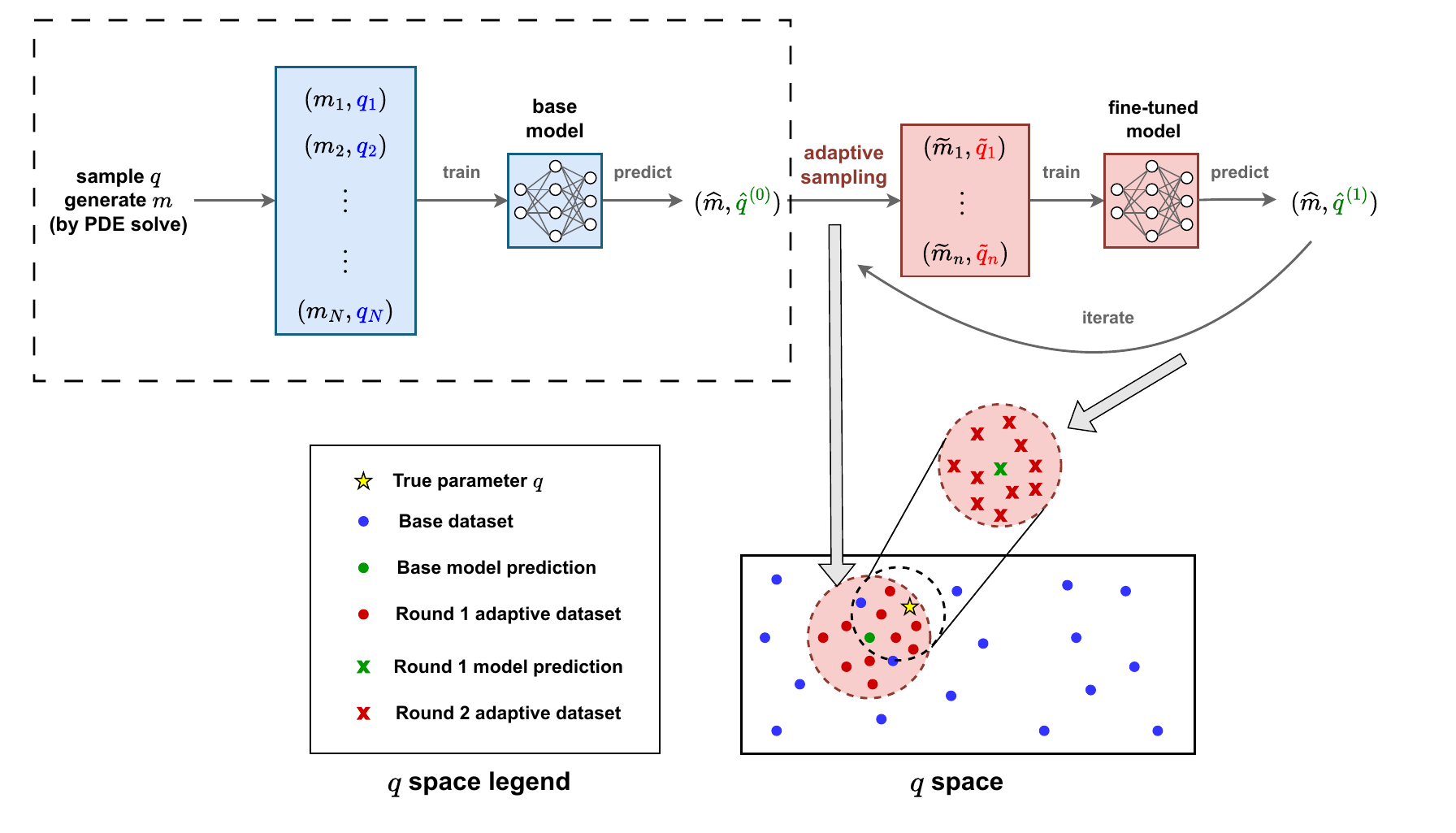}
\caption{Schematic of the instance-wise adaptive sampling method. The upper-left portion of the diagram in the dashed box depicts the typical machine learning approach to inverse problems, resulting in a base model for the inverse operator and its prediction of the unknown parameter corresponding to a given measurement instance. In the adaptive sampling method, the base model and its prediction are iteratively refined, as depicted in the upper-right portion of the diagram. It is important to note that these iterative refinements are specifically tailored to the given measurement instance. The bottom of the figure shows the progression of the method in the parameter space.}
\label{FIG:Schematic}
\end{figure}

\begin{enumerate}
    \item Train with a small amount of base data to obtain a crude base model $\mathcal{NN}_{\theta_0}$, where $\theta_0$ denotes the learned model weights, which should not be confused with the physical parameters we aim to reconstruct.

    \item Given a new measurement instance $\widehat{m}$, apply the base model to obtain an initial estimate $\widehat{q}^{(0)} = \mathcal{NN}_{\theta_0}(\widehat{m})$ of the associated parameter.
    
    \item Project $\widehat{q}^{(0)}$ onto the parameter manifold $\mathcal{M}$, yielding the closest point on $\mathcal{M}$ under a suitable distance metric. This step ensures that subsequent sampling is constrained to the prior-informed parameter space.

    \item Generate a new adaptive dataset by randomly sampling from the parameter manifold $\mathcal{M}$ around the projection of $\widehat{q}^{(0)}$. Fine-tune the current model on this local dataset (possibly together with some base data) to update its model weights to $\theta_1$, and apply the new model to the measurement $\widehat{m}$ to obtain an improved estimate $\widehat{q}^{(1)}$ of the parameter.
    
    \item Repeat the above projection (step 2), sampling (step 3), and refinement (step 4) for a number of rounds or until convergence, producing increasingly accurate estimates $\widehat{q}^{(1)}, \widehat{q}^{(2)}, \dots$ of the desired parameter.
\end{enumerate}

This procedure is \textit{instance-wise} in the sense that the data generated in later rounds is tailored to the specific test measurement $\widehat{m}$ and varies across different instances. A schematic illustration of the method is shown in Figure~\ref{FIG:Schematic}, and the complete procedure is summarized in Algorithm~\ref{ALGO:Adaptive Sampling}. Several components of this high-level workflow will be discussed in more detail later in the paper. In particular:
\begin{itemize}
    \item Note that the projection onto the manifold $\mathcal{M}$ in step 3 (line~\ref{alg:line_proj} in Algorithm~\ref{ALGO:Adaptive Sampling}) and the random perturbation of the parameter on $\mathcal{M}$ in step 4 (line~\ref{alg:line_perturb} in Algorithm~\ref{ALGO:Adaptive Sampling}) depend on specific prior knowledge of the data manifold. These procedures will be described in more detail in the next section, based on the two types of priors considered in this work.

    \item Section~\ref{SEC:Numerical} provides further details on the implementation and hyperparameters used in constructing the adaptive dataset in step 4 (line~\ref{alg:adaptive_dataset}).

    \item Details on the fine-tuning training process in step 4 (line~\ref{alg:fine-tune}) will be discussed in Section~\ref{SEC:Numerical} as well.

    \item Stopping criteria for the algorithm in step 5 (line~\ref{alg:stopping_criterion}) will also be discussed in Section~\ref{SEC:Numerical}.
\end{itemize}

\begin{algorithm}[!htb]
\caption{Adaptive Sampling for Inverse Problems}
\label{ALGO:Adaptive Sampling}
\textbf{Given:} Forward operator $\mathcal{F}$, parameter manifold $\mathcal{M}$, base model $\mathcal{NN}_{\theta_0}$ approximating $\mathcal{F}^{-1}$, base model dataset $\mathcal{D}_\text{base model}$\\
\textbf{Input:} Measurement $\widehat{m}$\\
\textbf{Hyperparameters:} $N_\text{adapt}$
\begin{algorithmic}[1]
    \State $\widehat{q}^{(0)} = \mathcal{NN}_{\theta_0}(\widehat{m})$  \Comment{Prediction from initial base model}
    \State $t = 0$
    \While{stopping criterion not met} \label{alg:stopping_criterion}
        \State{$\widehat{q}^{(t)} \gets \text{Projection of } \widehat{q}^{(t)} \text{ onto parameter manifold }\mathcal{M}$} \label{alg:line_proj}
        \For{$i = 1,2,\dots,N_\text{adapt}$} \Comment{Generate adaptive dataset}
            \State Randomly perturb $\widehat{q}^{(t)}$ on $\mathcal{M}$ to obtain $\widetilde{q}_i$ \label{alg:line_perturb}
            \State $\widetilde{m}_i = \mathcal{F}(\widetilde{q}_i)$
        \EndFor
        \State Form an adaptive dataset $\mathcal{D}_t$ from $\{(\widetilde{m}_i, \widetilde{q}_i)\}_{i=1}^{N_\text{adapt}}$ and possibly some elements of $\mathcal{D}_\text{base model}$ \label{alg:adaptive_dataset}
        \State Update the model weights of $\mathcal{NN}$ from $\theta_{t}$ to $\theta_{t+1}$ by fine-tuning on the adaptive dataset $\mathcal{D}_t$ \label{alg:fine-tune}
        \State $\widehat{q}^{(t+1)} = \mathcal{NN}_{\theta_{t+1}}(\widehat{m})$ \Comment{Refined prediction}
        \State $t \gets t+1$
    \EndWhile
    \State \textbf{return} $\widehat{q}^{(t)}$
\end{algorithmic}
\end{algorithm}

\subsection{Analogy with Inference-Time Compute}
The proposed method is part of a broader trend in machine learning that shifts more computation to the inference stage, a direction that has gained significant traction in the context of large language models (LLMs) \cite{snell2024scaling, liu2025can, openai2024o1}. For LLMs, inference-time computation typically falls into two main categories, as illustrated in \cite[Figure 5]{snell2024scaling}: (1) parallel sampling \cite{Brown-etal-arXiv24, StKaNa-arXiv24}, and (2) sequential revision \cite{madaan2024self, qu2025recursive, Welleck-etal-arXiv22}, with recent work also exploring hybrids of both approaches.

In parallel sampling, the LLM is queried multiple times with the same prompt, producing diverse outputs. A separate verifier then selects the best response. In tasks such as code generation and mathematical reasoning, the verifier often takes the form of unit tests or formal proof assistants. In the context of our inverse problems, we already have a good verifier due to the nature of the problem: the discrepancy measure $\mathcal{L}$ in \eqref{eq:general_loss}, which quantifies how well a reconstructed parameter matches the observed measurement under the forward model.

On the other hand, in sequential revision, the LLM first generates an initial solution and then iteratively refines its answer. While our instance-wise adaptive sampling method does not precisely align with existing LLM inference-time paradigms, it shares structural similarities with the sequential revision framework. In what follows, we draw a concrete analogy using Self-Refine approach introduced in \cite{madaan2024self} as a representative example.

For ease of explanation, we reproduce the pseudocode of Self-Refine from \cite[Algorithm 1]{madaan2024self} as Algorithm~\ref{ALGO:Self-Refine}. The method begins with a preliminary generation of the answer from the LLM, followed by a feedback step in which the same model critiques the answer, and a refinement step in which the model incorporates the feedback to produce an improved version. This process is repeated for multiple rounds. Few-shot examples are used in the prompt to guide the model during generation, feedback, and refinement, denoted in Algorithm~\ref{ALGO:Self-Refine} by $p_\text{gen}$, $p_\text{fb}$, and $p_\text{refine}$ respectively, with $||$ indicating prompt concatenation.

\begin{algorithm}
\caption{LLM Self-Refine \cite[Algorithm 1]{madaan2024self}}
\label{ALGO:Self-Refine}
\textbf{Given:} User input $x$, LLM model $\mathcal{P}$, few-shot prompts $p_\text{gen}$, $p_{\text{fb}}$, and $p_{\text{refine}}$
\begin{algorithmic}[1]
    \State $y_0 = \mathcal{P}(p_\text{gen} \sep x)$
    \For{$t = 0,1,\dots,T-1$}
        \State $fb_t = \mathcal{P}(p_\text{fb} \sep x \sep y_t)$ \Comment{Feedback}
        \State $y_{t+1} = \mathcal{P}(p_\text{refine} \sep x \sep y_0 \sep fb_0 \sep \dots \sep y_t \sep fb_t)$ \Comment{Refine}
    \EndFor
    \State \textbf{return} $y_T$
\end{algorithmic}
\end{algorithm}

Table~\ref{TAB:LLM Comparison} summarizes the analogy between Self-Refine approach for LLM and our adaptive sampling method for inverse problems. A key parallel lies in the iterative refinement structure: in both cases, the model begins with an initial prediction and improves it over successive rounds using feedback. In the LLM setting, feedback is explicitly generated text based on the input and the model's prior output. In contrast, our method constructs an adaptive dataset by perturbing the current estimate, which serves as implicit feedback used to fine-tune the model. Unlike Self-Refine, we do not explicitly evaluate or critique intermediate outputs; rather, refinement emerges through localized resampling and model updating, informed by prior knowledge of the parameter space and access to a forward operator.

The comparison also reveals some differences. LLMs typically operate with frozen model weights during inference, leveraging prompt engineering and in-context learning to refine outputs. In contrast, our model is explicitly fine-tuned at inference time using newly collected, instance-specific data. This distinction reflects differing priorities: while LLMs prioritize zero-shot generality, our method is tailored for high-accuracy instance-wise reconstruction in structured scientific domains.

More broadly, this analogy also suggests that other inference-time strategies developed for LLMs could inspire new adaptive sampling techniques for inverse problems.

\begin{table}[!htb]
    \centering
    \renewcommand{\arraystretch}{1.5}
    \begin{tabular}{p{8em} p{13em} p{15.5em}}
    \toprule
         & \textbf{LLM Self-Refine} & \makecell[l]{\textbf{Adaptive Sampling for} \\\textbf{Inverse Problems}}\\
    \midrule
        \textbf{Model Input} & User input $x$ & Measurement $m$\\
    \midrule
        \textbf{Model Output} & Response $y$ & Parameter $q$ \\
    \midrule
        \textbf{Feedback \quad Process} & Use the model, input $x$, and latest output $y_t$ to generate the feedback $fb_t$ & Generate adaptive dataset of perturbations around the latest output $\widehat{q}^{(t)}$ and interpret the result as the feedback $fb_t$ \\
    \midrule
        \textbf{Refinement Process} & Use the model, input $x$, past outputs $y_0, \dots, y_t$, and corresponding feedbacks $fb_0, \dots, fb_t$ to obtain refined output $y_{t+1}$ & Fine-tune the model on the adaptive dataset (i.e., feedback) $fb_t$, then predict on input $m$ to obtain refined output $\widehat{q}^{(t+1)}$ \\
    \bottomrule
    \end{tabular}
    \caption{Comparison between LLM Self-Refine \cite{madaan2024self} and our adaptive sampling method for inverse problems.}
    \label{TAB:LLM Comparison}
\end{table}

\subsection{Other Related Works}
There has also been related work in the applied mathematics and scientific computing communities. Perhaps the closest to our setting is \cite{tatsuoka2025multi}, which introduces an instance-wise adaptive refinement method in the context of Bayesian inverse problems. Their objective is to characterize the full posterior distribution of the parameter, which leads them to focus on low-dimensional parameter spaces (one- or two-dimensional). Their method also involves only two sampling levels, whereas ours allows multiple rounds of refinement for high-dimensional parameters. Overall, the two approaches share some spirit at a high level but are not directly comparable.

The remaining related work can be broadly divided into two main categories. One line studies other machine learning approaches to inverse problems. For example, \cite{melia2025multi} proposes a neural network architecture for the multi-frequency inverse scattering problem, first constructing a low-frequency approximation from the lowest-frequency measurement and then iteratively refining it with higher-frequency data. Another recent work \cite{jiang2024reinforced} uses reinforcement learning to adaptively choose sensor locations and incident frequencies, highlighting the benefits of adaptivity in frequency design for inverse scattering. By contrast, we focus on the single-frequency case, and our progressive refinement is not frequency-based. Despite the difference, the analogy in their refinement structure suggests an interesting future direction: frequency-based adaptive sampling, where early rounds reconstruct low-frequency components that are subsequently refined into higher-frequency approximations.

The second category applies sequential adaptive sampling to settings outside inverse problems, such as adaptive collocation point selection for physics-informed neural networks \cite{lu2021deepxde, wu2023comprehensive} and adaptive proposal construction for rare event probability estimation \cite{tong2023large}. Although the applications and objectives differ, these methods share a structural similarity with ours: at each round, random samples are drawn adaptively based on the current state, used to update the state, and repeated over multiple rounds. The successes in these areas highlight the versatility of sequential adaptive sampling and suggest that it could be fruitfully explored in still more domains.

\section{Example Problem: Inverse Scattering}

Note that the methodology put forward in the previous section is a general one that can be applied in principle to various inverse problems. In the numerical experiments of this paper, we demonstrate the effectiveness of the method by applying it to the inverse scattering problem. In this inverse problem, one seeks to reconstruct properties of an object by sending incident waves at the object and measuring the scattered waves at receivers. 

More specifically, we consider the inverse acoustic scattering problem in two dimensions, where the goal is to reconstruct the scattering potential, i.e., the relative refractive index, $q(\bx)$ of a medium, defined as a function on $\bbR^2$. The scattering potential is related to the spatially varying wave speed $c(\bx)$ by the relation $q(\bx) = c_0^2 / c^2(\bx) - 1$, where $c_0 \equiv 1$ denotes the normalized wave speed in free space. Consequently, recovering $q(\bx)$ enables the determination of the wave speed distribution within the object, providing insights into its physical properties. We assume that the scatterer is contained within a domain $\Omega = [-\pi/2, \pi/2]^2$, so that by definition $q(\bx)$ is compactly supported in $\Omega$. With a slight abuse of notation, we will use $q$ to refer both to the function defined on $\bbR^2$ and its restriction to $\Omega$. 

To model wave propagation in this setting, we adopt a time-harmonic formulation, where the response to monochromatic sources is governed by the Helmholtz equation. Specifically, sending an incoming plane wave $u^{\text{inc}}(\bx) = \exp(ik\bx\cdot \bd)$ with wavenumber $k$ and direction $\bd \in S^1$ results in a scattered wave $u^{\text{scat}}(\bx)$. Here, $i$ denotes the imaginary unit. The scattered wave is defined so that the total wave $u(\bx) = u^{\text{inc}}(\bx) + u^{\text{scat}}(\bx)$ satisfies the following Helmholtz problem:
\begin{equation}
    \left\{
    \begin{aligned}
        \Delta u(\bx) + k^2(1+q(\bx)) u(\bx) &= 0, \qquad \text{in }\bbR^2,\\
        \lim_{r\rightarrow\infty} \sqrt{r}\left(\frac{\partial u^{\text{scat}}}{\partial r} - iku^{\text{scat}}\right) &= 0,\qquad r = |\bx|.
    \end{aligned}
    \right.
\end{equation}

\begin{figure}[!htb]
\centering
\begin{subfigure}{0.35\textwidth}
    \centering
    \includegraphics[width=\textwidth]{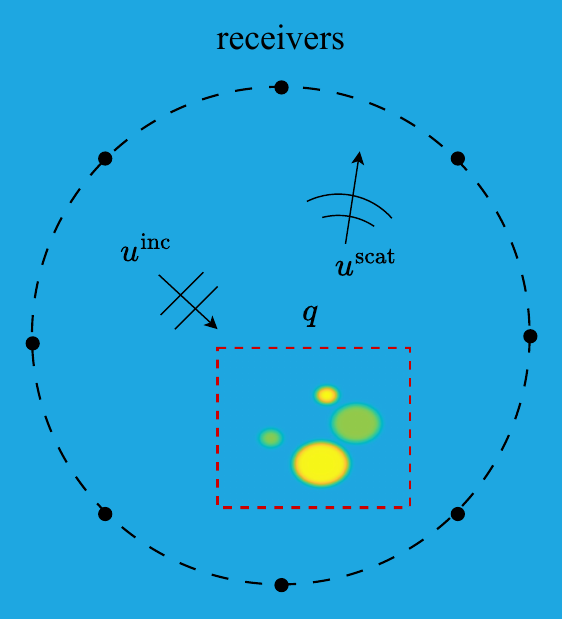}
\end{subfigure}\hspace{20pt}
\begin{subfigure}{0.5\textwidth}
    \centering
    \includegraphics[width=\textwidth]{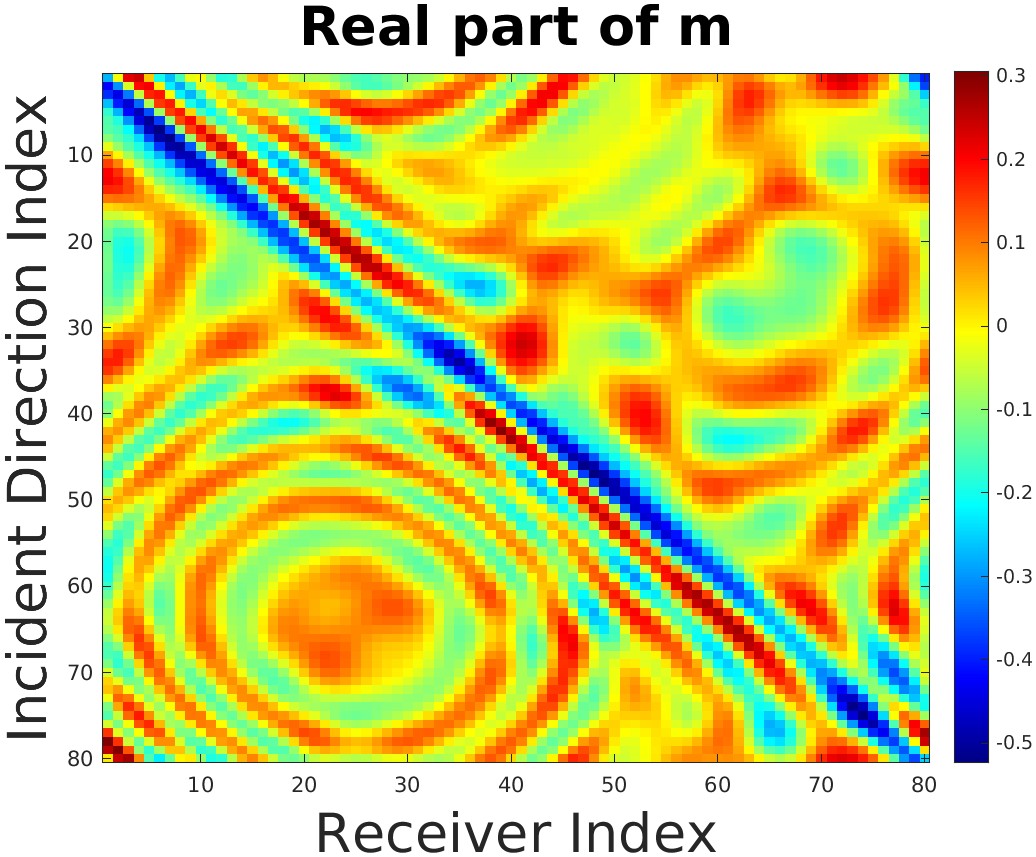}
\end{subfigure}
 \caption{A schematic of the inverse scattering problem. \textbf{Left:} Illustration of the experimental setup, in which an incident wave scatters off the medium and is detected at $N_t$ receivers. A total of $N_d$ incident waves, sent from different directions, are used to obtain the full measurement. \textbf{Right:} The resulting measurement matrix $m\in \bbC^{N_d\times N_t}$. For visualization purposes, only the real part of $m$ is displayed.}
\label{FIG:Inverse Scattering}
\end{figure}

Let $N_t$ denote the number of receivers and $\{\bx_\ell\}_{\ell=1}^{N_t}$ the receiver locations, which are typically located far away from the domain $\Omega$. Assume also that we are able to measure data at the receivers for multiple incident directions, denoted by $\{\bd_j\}_{j=1}^{N_d}$. 
The forward operator $\mathcal{F}_k: \mathcal{Q}\rightarrow \bbC^{N_d\times N_t}$ of this inverse problem is then defined as 
\begin{equation}
    \mathcal{F}_k(q) = m,
\end{equation}
where the $(j,\ell)$ entry of the matrix $m\in \bbC^{N_d\times N_t}$ is given by $u^{\text{scat}}_{k, \bd_j}(\bx_\ell)$. Here $\mathcal{Q}$ is the space of smooth functions on $\bbR^2$ supported on the domain $\Omega$. See Figure~\ref{FIG:Inverse Scattering} for a schematic of this inverse scattering problem.

To discretize the parameter space $\mathcal{Q}$ and enable a numerical formulation of the inverse problem, we begin by noting that any compactly supported function on the considered domain $\{\bx = (x, y) \in \Omega\}$ can be represented using the sine basis $\{\sin(i(x+\pi/2))\sin(j(y+\pi/2))\}_{i,j\ge 1}$. Based on this, we define the finite-dimensional subspace
$$
\mathcal{Q}_N \coloneq \operatorname{span}\Big\{\sin(i(x+\pi/2))\sin(j(y+\pi/2))\Big\}_{1\le i,j\le N}.
$$
Here, $\operatorname{span}$ denotes the set of all linear combinations of the basis functions specified. We then choose a truncation level $N_3$ as an integer roughly on the order of the wavenumber $k$, and formulate the inverse problem as the following optimization task: 
\begin{equation}
\label{eq:loss}
    \min_{q\in \mathcal{Q}_{N_3}} \|\mathcal{F}_k(q) - m\|^2,
\end{equation}
where $m$ is the given measurement data. It is important to note that the number of basis functions in $\mathcal{Q}_{N_3}$ may be significantly smaller than that needed to fully resolve the fine-scale features of the ground-truth field $q^*$. This deliberate restriction of the search space helps mitigate the ill-posedness inherent in the inverse scattering problem, which is fundamentally constrained by the Heisenberg uncertainty principle~\cite{chen1997inverse}. A similar strategy was adopted in~\cite{borges2017high,askham2024reconstructing} to address the same challenge. Note that while the restricted search space $\mathcal{Q}_{N_3}$ can be seen as prior information on $q^*$, in our setup, the prior manifold $\mathcal{M}$ is a separate space and is not necessarily related to the smoothness of $q^*$. Two examples of the prior manifold $\mathcal{M}$ are discussed in the next subsection.

\subsection{Prior Knowledge of the Parameter}
\label{sec:prior}
In this subsection, we go into more detail on the space of parameters. First, we fix a large $N$ ($128$ in our experiments), and assume that the true field $q^*$ lies in the space $\mathcal{Q}_{N}$. In the notation of Section~\ref{SEC:Methodology}, this means that the dimension $N_2$ of the parameter ambient space is $N^2 = 128^2$.

As mentioned earlier, prior knowledge may indicate that $q^*$ lies on a specific manifold $\mathcal{M} \subset \mathcal{Q}_{N}$. In this paper, we investigate two such manifolds, referred to as the \textit{disk prior} and the \textit{Fourier prior}.

\subsubsection{Disk prior}
\label{SEC:disk prior}
In the disk prior setting, the prior assumption on the parameter $q^*$ is that it is made up of a collection of disjoint disks with constant amplitude; a typical example of such a field can be found at the top of Figure~\ref{FIG:Circle visualization}. The dimension of the prior manifold $\mathcal{M}$ is then determined by the maximum number of disks $N_{\text{disk}}$, and each data on the manifold $\mathcal{M}$ is determined by the number of disks, the location, size, and the constant amplitude of each disk. For a more precise description, see Appendix \ref{APP:disk prior}.

In this setting, the projection onto $\mathcal{M}$ in line~\ref{alg:line_proj} and the local perturbation on $\mathcal{M}$ in line~\ref{alg:line_perturb} of Algorithm \ref{ALGO:Adaptive Sampling} can be implemented as follows. Given the prediction $\widehat{q}^{(t)}$ of the current neural network model $\mathcal{NN}_{\theta_t}$, we utilize the phase-coding method~\cite{atherton1999size}, implemented in the \texttt{imfindcircles} function in MATLAB, to detect all the possible disks $\mathcal{D}_1,\dots,\mathcal{D}_n$ in $\widehat{q}^{(t)}$.  The field $\widehat{q}^{(t)}$ is then averaged over each disk $\mathcal{D}_i$ to obtain an associated amplitude $a_i$. The collection of disks $\mathcal{D}_1,\dots,\mathcal{D}_n$ together with their amplitudes $a_1,\dots,a_n$ corresponds to the projection onto the prior manifold in line~\ref{alg:line_proj} of Algorithm \ref{ALGO:Adaptive Sampling}. We then sample around $\widehat{q}^{(t)}$ on the parameter manifold $\mathcal{M}$ by randomly perturbing the centers, radii, and amplitudes of each of the disks. 

It should be pointed out that due to the inherent difficulty of the disk detection task and the behavior of the \texttt{imfindcircles} function, the detected disks $\mathcal{D}_1, \dots, \mathcal{D}_n$ may overlap. In such cases, the resulting configuration does not lie exactly on the parameter manifold $\mathcal{M}$. Nonetheless, it can still be viewed as an approximate projection onto $\mathcal{M}$. See the bottom-middle plot in Figure~\ref{FIG:Circle visualization} for an example of this phenomenon.

\subsubsection{Fourier prior}
\label{SEC:Fourier prior}
The first type of prior incorporates strong structural knowledge about the underlying field. We now turn to a much more generic prior based on the Fourier coefficients. In the Fourier prior setting, the parameter field $q^*$ is assumed to be bandlimited to a small fixed number $N_F$ of Fourier modes, defined with respect to a smaller domain $\Omega' := [-\pi/2 + \varepsilon, \pi/2 - \varepsilon]^2$. As part of the prior knowledge, $q^*$ is taken to vanish outside $\Omega'$, so that its support is effectively contained within this interior region.
See Appendix \ref{APP:Fourier prior} for a more precise description of the prior. The number $N_F$ controls the dimension $N_1$ of the prior manifold $\mathcal{M}$ (specifically, $N_1$ is proportional to $N_F^2$), and in our numerical experiments $N_F$ is chosen as $3$ or $4$. A typical example of such a field for $N_F=3$ can be found at the top of Figure~\ref{FIG:Fourier visualization}.

Given the structure of the prior manifold $\mathcal{M}$ in this setting, the projection onto $\mathcal{M}$ in line~\ref{alg:line_proj} and the local perturbation on $\mathcal{M}$ in line~\ref{alg:line_perturb} of Algorithm \ref{ALGO:Adaptive Sampling} can be implemented as follows. First, the projection of the field $\widehat{q}^{(t)}$ onto the parameter manifold $\mathcal{M}$ is performed by computing the coefficients of the first $N_F$ Fourier modes of the restriction of the field to the smaller domain $\Omega'$. To perform local sampling around this projection on the manifold $\mathcal{M}$, each Fourier coefficient is then perturbed by zero-mean random noise of a certain standard deviation. Note that the standard deviation should be positively correlated with the error of the current field estimate $\widehat{q}^{(t)}$ to the true field, which we estimate with the help of the validation set used to train the current neural network model. Again, we refer the reader to Appendix \ref{APP:Fourier prior} for a more detailed explanation of how the standard deviations of the perturbations are calculated. It is worth noting that the perturbation standard deviation tends to decrease with successive rounds, reflecting the increasing accuracy of $\widehat{q}^{(t)}$.

\section{Numerical Results}
\label{SEC:Numerical}

In the following experiments, we fix the wavenumber $k$ at $15$. Incident waves are sent from 80 equally spaced directions, and the scattered field is measured by 80 equally spaced receivers placed along the boundary of a circle of radius 10. This setup corresponds to $N_d = 80$ incident directions and $N_t = 80$ receivers. In the minimization formulation \eqref{eq:loss} of the inverse problem, we set $N_3 = k = 15$, following the principle discussed above, which amounts to optimizing the $N_3^2 = 225$ basis coefficients in $\mathcal{Q}_{N_3}$. 

The network architecture consists of $L$ convolutional layers followed by fully-connected layers. The input to the network is an $N_d \times N_t$ complex-valued scattering measurement, represented as two channels (real and imaginary parts).
Each convolutional layer $ \ell \in \{1, \dots, L\} $ with $N_c$ channels employs $K_\text{conv} \times K_\text{conv}$ kernels with periodic padding \(p\), followed by a ReLU activation and average pooling with kernel size $K_\text{pool}$ and stride $s$. The output of the convolutional blocks is flattened and subsequently processed by a sequence of fully-connected layers. Each of these layers applies a ReLU activation, and their respective output dimensions are specified by the tuple $\mathbf{d}_\text{fc}$. Specific hyperparameters for networks used in two distinct priors are detailed in Table~\ref{TAB:CNN parameters}.

\begin{table}[!htb]
    \centering
    \begin{tabular}{c|cccccc}
    \toprule
         Prior & $(N_d, N_t)$ & $L$ & $N_c$ & $(K_\text{conv}, p)$ & $(K_\text{pool}, s)$ & $\mathbf{d}_\text{fc}$\\
    \midrule
        Disk & (80, 80) & 2 & 64 & (5, 2) & (2, 2) & (512, 256, 225)\\
        Fourier & (80, 80) & 3 & 64 & (5, 2) & (2, 2) & (512, 256, 225)\\
    \bottomrule
    \end{tabular}
    \caption{Parameters of the network architecture. The input is a 2-channel array of size $N_d \times N_t$. For the $L$ convolutional layers: $N_c$ is the number of channels, $K_\text{conv}$ is the kernel width, and $p$ is the periodic padding width. For average pooling: $K_\text{pool}$ is the kernel width and \(s\) is the stride. $\mathbf{d}_\text{fc}$ specifies the sequence of output dimensions for the fully-connected layers.}
    \label{TAB:CNN parameters} 
\end{table}

We choose to implement the construction of the adaptive datasets in line~\ref{alg:adaptive_dataset} of Algorithm~\ref{ALGO:Adaptive Sampling} by combining the adaptively sampled local dataset of size $N_\text{adapt}$ with the $N_\text{base}$ elements in the base model dataset whose parameter fields are closest to the current prediction, measured in terms of the $\ell^2$ norm in $\bbR^{N_3^2}$. We note that this way of combining local data and base data, along with the specific choices of $N_\text{adapt}$ and $N_\text{base}$ in Tables~\ref{TAB:Circle dataset size parameters} and~\ref{TAB:Fourier dataset size parameters}, are current design choices; alternative configurations can certainly be explored.

The training of the network is performed using stochastic gradient descent with momentum 0.9 and batch size $100$ on a normalized dataset, and the loss function is the mean squared $\ell^2$ error in $\bbR^{N_3^2}$. We use a learning rate of $0.1$ to train the base models and a smaller learning rate of $0.01$ to train the fine-tuned models during the adaptive rounds. To prevent overfitting, we use early stopping with a validation set constructed in a similar way to the training set. In particular, during the adaptive rounds, the validation set is a combination of adaptively sampled data and base model data, where the ratio of the two dataset sizes is the same as in the training set.

The reconstruction error of $\widehat{q}$ is measured by the relative $\ell^2$ error in $\bbR^{N_3^2}$ between the predicted coefficient vector and the corresponding coefficient vector for the ground-truth field $q^*$. The test error $\varepsilon_\text{rel}$ is then defined as the average relative error over a test set.

To systematically quantify the improvement brought by adaptive sampling for data construction within the learning-based method over its non-adaptive counterpart, as well as over traditional optimization-based approaches, we run the adaptive method for a fixed number of rounds $N_\text{round}$ on all test data, rather than using a data-dependent stopping criterion as in Algorithm~\ref{ALGO:Adaptive Sampling}.  This allows us to track the average relative error $\varepsilon_\text{rel}$ after each round. The value of $N_\text{round}$ is chosen so that the average error across all test data plateaus, as illustrated in the left panels of Figures~\ref{FIG:Circle curves} and~\ref{FIG:Fourier curves}. Note that in practice, we do not have access to the reconstruction error because we do not know the ground-truth field $q^*$. Instead, we can track the measurement error, i.e., the relative $\ell^2$ error in $\bbR^{N_d\times N_t}$ between the measurement associated to the reconstructed parameter $\widehat{q}$ and the given measurement. A practical stopping criterion is to terminate when the measurement error begins to plateau.

\subsection{Disk prior}  
\begin{figure}[!htb]
\centering
\includegraphics[width=\textwidth]{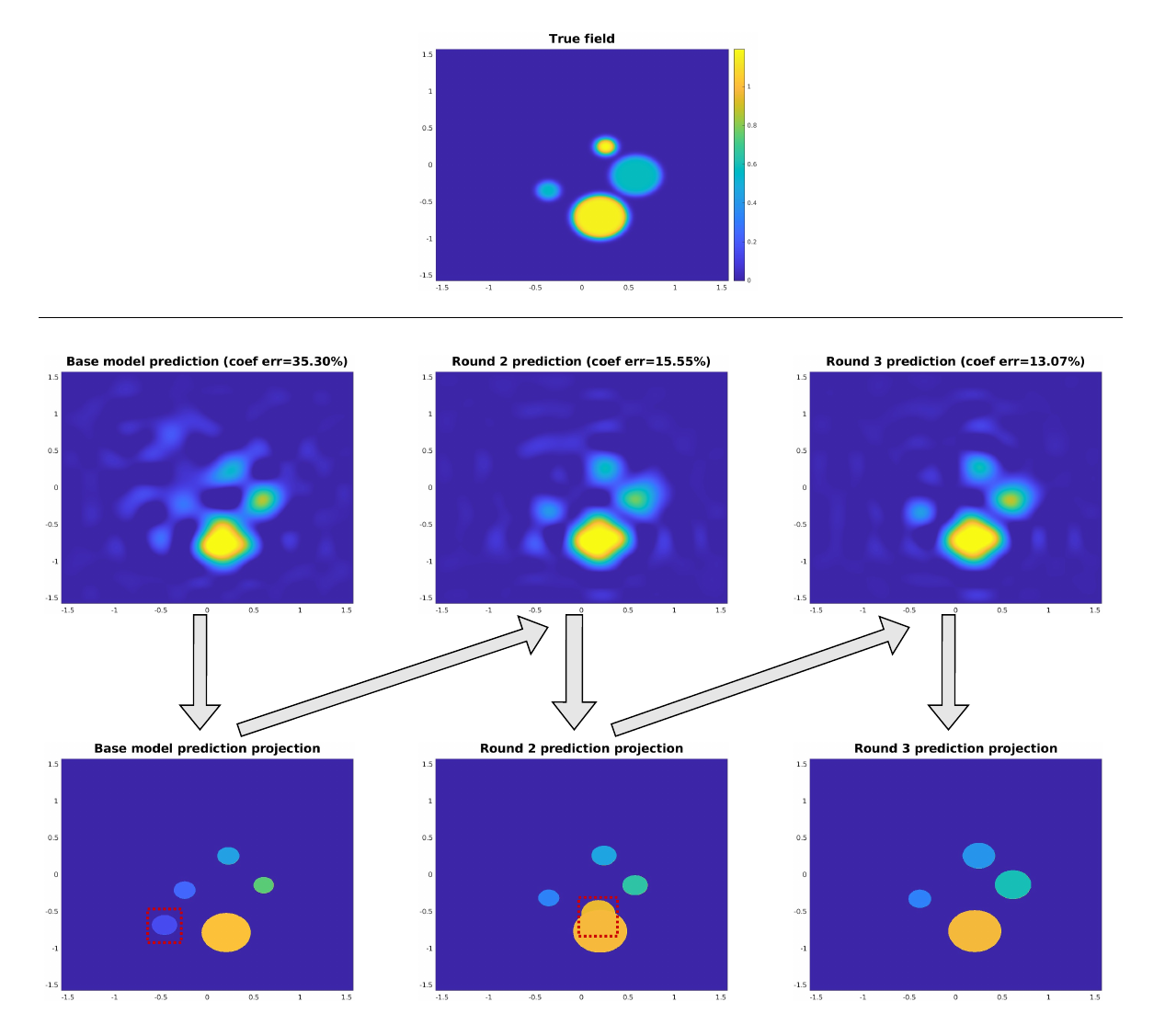}
\caption{Visualization of field progression for a test case under the $N_\text{disk} \in [4,6]$ disk prior setting. \textbf{Top:} Ground truth field. \textbf{Middle:} Predicted fields from the base model and subsequent refinement rounds. \textbf{Bottom:} Projections of the predicted fields onto the disk prior manifold. The projections from the base model and round 2 include extra disks (highlighted by red dashed squares) that are not present in the true field. Nevertheless, these errors are progressively corrected in later rounds.}
\label{FIG:Circle visualization}
\end{figure}

We consider two different settings for the problem with disk prior: $N_{\text{disk}}\in [1,3]$ and $N_{\text{disk}}\in [4,6]$, in which the number of disks is chosen uniformly at random from the indicated set. Recall that for a fixed $N_{\text{disk}}$, the dimension of the corresponding data manifold is $4N_{\text{disk}}$. The dataset size hyperparameters for our adaptive sampling method are laid out in Table~\ref{TAB:Circle dataset size parameters}. Figure~\ref{FIG:Circle visualization} depicts the progression of the reconstructed field in the adaptive sampling method for a specific test instance with $N_{\text{disk}}=4$. It is worth highlighting that the method is robust to errors in the base model prediction. Specifically, note that the projection of the base model prediction onto the disk prior manifold introduces an additional disk absent in the true field. Nevertheless, after several iterations, the method successfully corrects this initial mistake.

\begin{table}[!htb]
    \centering
    \begin{tabular}{c|ccc}
    \toprule
         Disk prior setting& $N_{\text{base model}}$ & $N_{\text{round}}$ & $(N_{\text{adapt}}, N_{\text{base}})$ \\ 
    \midrule
         $N_{\text{disk}}\in [1,3]$ & 1500 & 4 & (100, 200) \\ 
         $N_{\text{disk}}\in [4,6]$ & 5000 & 5 & (400, 800) \\
    \bottomrule
    \end{tabular}
    \caption{Dataset size hyperparameters for our adaptive sampling method in the disk prior setting. $N_{\text{base model}}$ denotes the size of the dataset $\mathcal{D}_\text{base model}$ used to train the initial base model. $N_{\text{round}}$ is the total number of adaptive rounds performed. In each round, the model is trained on a dataset consisting of $N_{\text{adapt}}$ adaptively generated local samples together with $N_{\text{base}}$ of the nearest samples from the dataset $\mathcal{D}_\text{base model}$.}
    \label{TAB:Circle dataset size parameters}
\end{table}

\begin{figure}[!htb]
\centering
\begin{subfigure}{\textwidth}
\begin{subfigure}{0.49\textwidth}
    \centering
    \includegraphics[width=\textwidth]{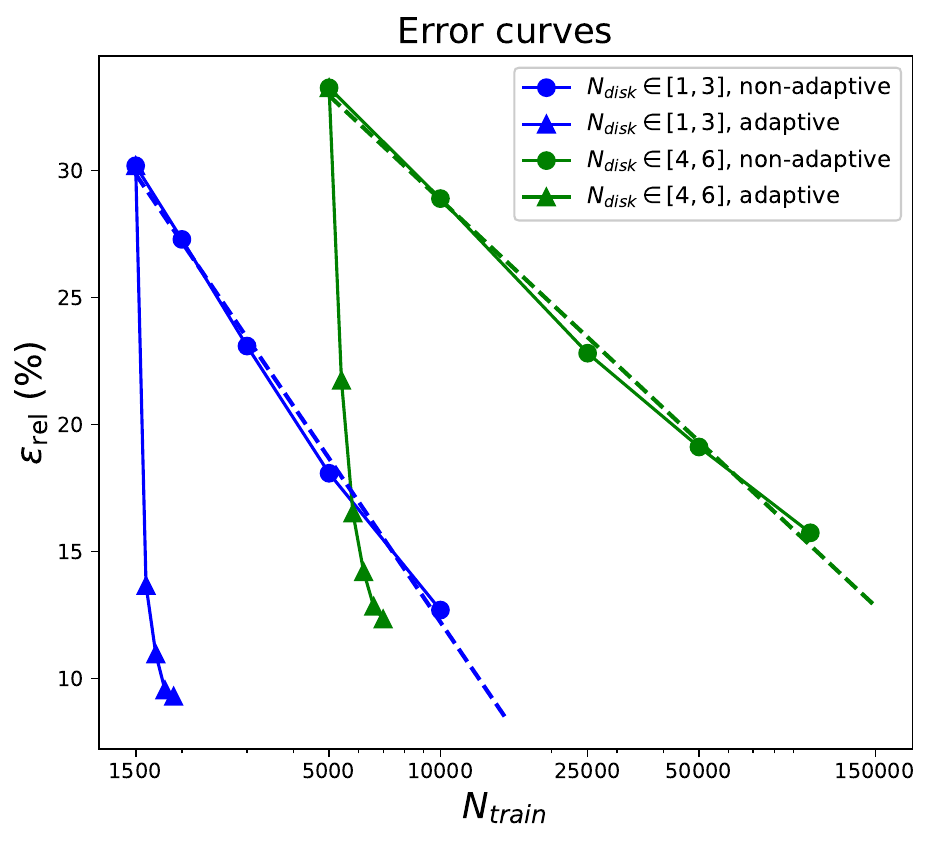}
\end{subfigure}
\hfill
\begin{subfigure}{0.49\textwidth}
    \centering
    \includegraphics[width=\textwidth]{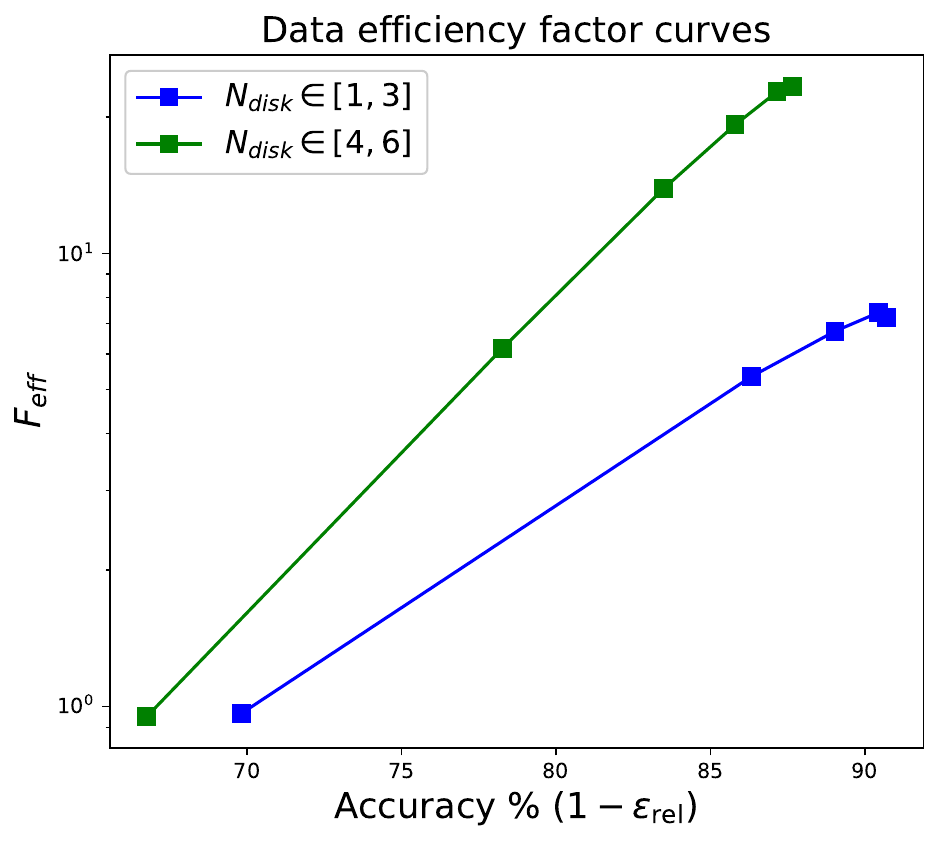}
\end{subfigure}
\end{subfigure}
 \caption{Data efficiency comparison between adaptive and non-adaptive training in the disk prior setting. \textbf{Left:} Average relative error $\varepsilon_\text{rel}$ versus total training dataset size. Dashed lines show a log-linear fit for the non-adaptive method across varying dataset sizes. \textbf{Right:} Data efficiency factor of the adaptive method, defined as the ratio of non-adaptive to adaptive dataset sizes required to reach the same error level, plotted as a function of target accuracy $1 - \varepsilon_\text{rel}$.}
\label{FIG:Circle curves}
\end{figure}

A comparison of the data scaling behavior between the standard non-adaptive one-shot training method and our adaptive sampling method is shown in the left plot of Figure~\ref{FIG:Circle curves}. For the adaptive method, we record the average relative error $\varepsilon_\text{rel}$ after each round of training. For the non-adaptive method, we train models with varying sizes of training data and measure the corresponding average relative errors. A linear regression (shown as a dashed line) is then performed between the error $\varepsilon_\text{rel}$ and the logarithm of the training set size. The fitted curve closely matches the actual data points, which suggests that the scaling behavior is well captured and allows us to estimate how many training samples would be required for the non-adaptive method to reach a given error level.

To quantify how much the adaptive sampling method reduces the number of needed training samples, we compute a \emph{data efficiency factor}, defined as the ratio between the estimated number of training samples needed by the non-adaptive method to reach a given error and the total number of samples used by the adaptive method at the same error level. These efficiency factors, plotted against the target accuracy $1 - \varepsilon_\text{rel}$, are shown in the right plot of Figure~\ref{FIG:Circle curves}.

As an illustrative example, consider the setting with $N_\text{disk} \in [4, 6]$. Suppose that we run the adaptive method for $5$ rounds. This means that for a single test case, our adaptive method requires generating $N_{\text{base model}} + N_\text{round}\cdot N_{\text{adapt}} = 5000 + 5\cdot 400 = 7000$ data samples, and the achieved relative error on average is  $12.3\%$.  According to the non-adaptive regression curve, achieving the same error would require approximately $163295$ training samples. Thus, the adaptive method yields a data efficiency factor of
\begin{equation*}
    F_{\text{eff}} = \frac{163295}{7000} \approx 23,
\end{equation*}
indicating a 23-fold reduction in the required data at that accuracy level.

In addition to the $N_{\text{disk}} \in [4,6]$ setting, Figure~\ref{FIG:Circle curves} also includes results for a simpler case with $N_{\text{disk}} \in [1,3]$. In the left plot, we observe that the slope of the non-adaptive dashed line shows a slower decay rate in the more complex setting ($N_{\text{disk}} \in [4,6]$) than in the simpler one, while the adaptive curves exhibit nearly identical slopes across both cases. This difference leads to two notable trends in the data efficiency curves in the right plot: (1) for a fixed prior, higher target accuracy gives rise to greater data efficiency; and (2) the efficiency curve for the more complex prior increases more rapidly than that for the simpler one. These observations highlight that the advantage of the adaptive method becomes more pronounced in more difficult inverse problems, either when higher accuracy is required or when the prior manifold is more complex.

We further compare the adaptive sampling approach with classical optimization-based methods; the results are summarized in Table~\ref{TAB:Disk GN}. Here, the adaptive model is trained on the adaptively constructed dataset, while the base model is trained on the initial dataset. The optimization problem \eqref{eq:loss} is solved using Gauss–Newton, following~\cite{borges2017high}. Since Gauss–Newton is sensitive to initialization, we report results with two strategies: nearest-neighbor initialization and base-model initialization. For nearest-neighbor initialization, we select the closest measurement in the base training set and use its associated parameter field as the initial guess. As shown in Table~\ref{TAB:Disk GN}, direct prediction from the adaptive model significantly reduces reconstruction error and outperforms Gauss–Newton even when initialized with the base-model prediction, indicating that the gain comes from adaptive data construction rather than post hoc optimization.    

\begin{table}[!htb]
    \centering
    \begin{tabular}{l|cc}
    \toprule
         & $N_{\text{disk}}\in [1,3]$ & $N_{\text{disk}}\in [4,6]$\\
    \midrule
        Direct prediction (base model) & 30.2\% & 33.3\% \\
        Direct prediction (adaptive model) & \textbf{9.3\%} & \textbf{12.0\%} \\
        Gauss--Newton (nearest-neighbor init.) & 29.4\% & 66.3\% \\
        Gauss--Newton (base-model init.) & 26.1\% & 29.1\% \\
    \bottomrule
    \end{tabular}
    \caption{Average reconstruction error on the test set under the disk prior, for direct prediction using the base and adaptive model (trained on the adaptively constructed dataset), and for Gauss--Newton refinement under two initialization strategies. Direct prediction from the adaptive model substantially outperforms both the base model and Gauss--Newton, even when the latter is initialized with the base-model prediction.}
    \label{TAB:Disk GN} 
\end{table}

\subsection{Fourier Prior}
\begin{figure}[!htb]
\centering
\includegraphics[width=\textwidth]{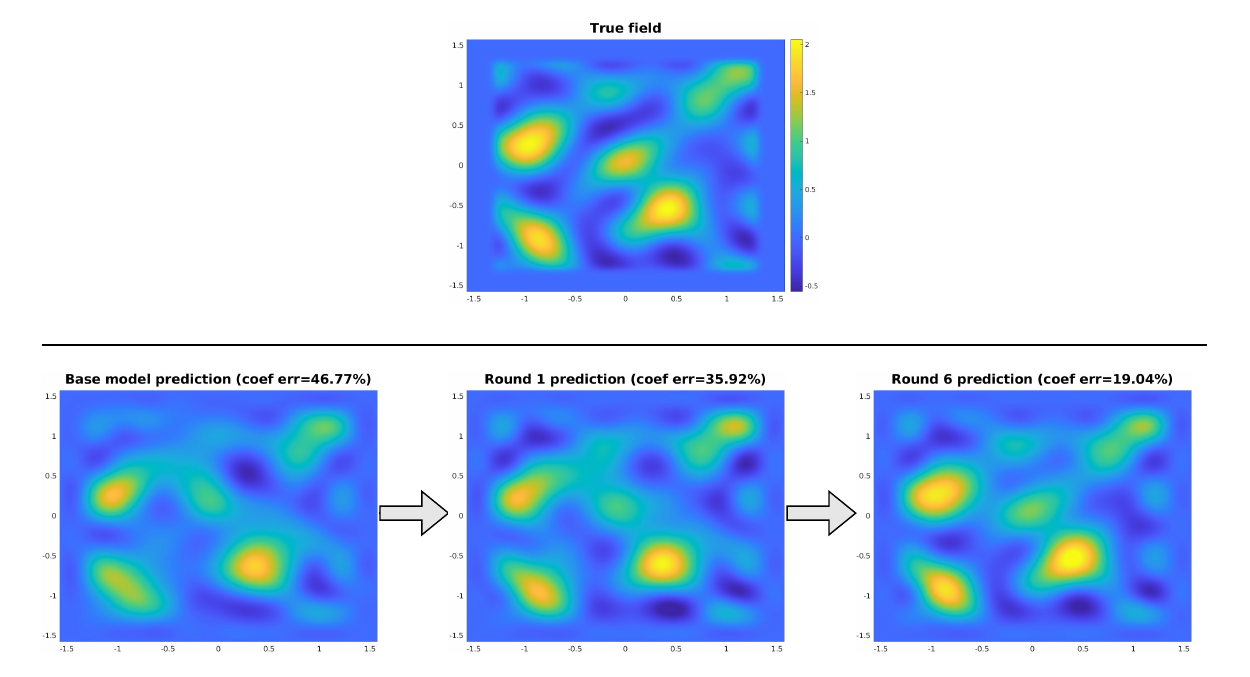}
\caption{Visualization of field progression for a test case under the $N_F=3$ Fourier prior setting. \textbf{Top:} Ground truth field. \textbf{Bottom:} Predicted fields from the base model and subsequent refinement rounds.}
\label{FIG:Fourier visualization}
\end{figure}

We consider two different settings for the problem with Fourier prior: $N_F=3$ and $N_F=4$. Recall that $N_F$ controls the number of Fourier modes and thus the dimension of the prior manifold $\mathcal{M}$, which scales proportionally to $N_F^2$. The dataset size hyperparameters used in our adaptive sampling method are listed in Table~\ref{TAB:Fourier dataset size parameters}, and an example of the progression of the reconstructed field along adaptive sampling is shown in Figure~\ref{FIG:Fourier visualization}.

\begin{table}[!htb]
    \centering
    \begin{tabular}{c|ccc}
    \toprule
         Fourier prior setting& $N_{\text{base model}}$ & $N_{\text{round}}$ & $(N_{\text{adapt}}, N_{\text{base}})$\\
    \midrule
        $N_F=3$ & 10000 & 6 & (500, 100) \\
        $N_F=4$ & 20000 & 7 & (1000, 100) \\
    \bottomrule
    \end{tabular}
    \caption{Dataset size hyperparameters for our adaptive sampling method in the Fourier prior setting. The definitions of $N_{\text{base model}}$, $N_{\text{round}}$, $N_{\text{adapt}}$, and $N_{\text{base}}$ are the same as in Table~\ref{TAB:Circle dataset size parameters}; see its caption for details.}
    \label{TAB:Fourier dataset size parameters} 
\end{table}

\begin{figure}[!htb]
\centering
\begin{subfigure}{\textwidth}
\begin{subfigure}{0.49\textwidth}
    \centering
    \includegraphics[width=\textwidth]{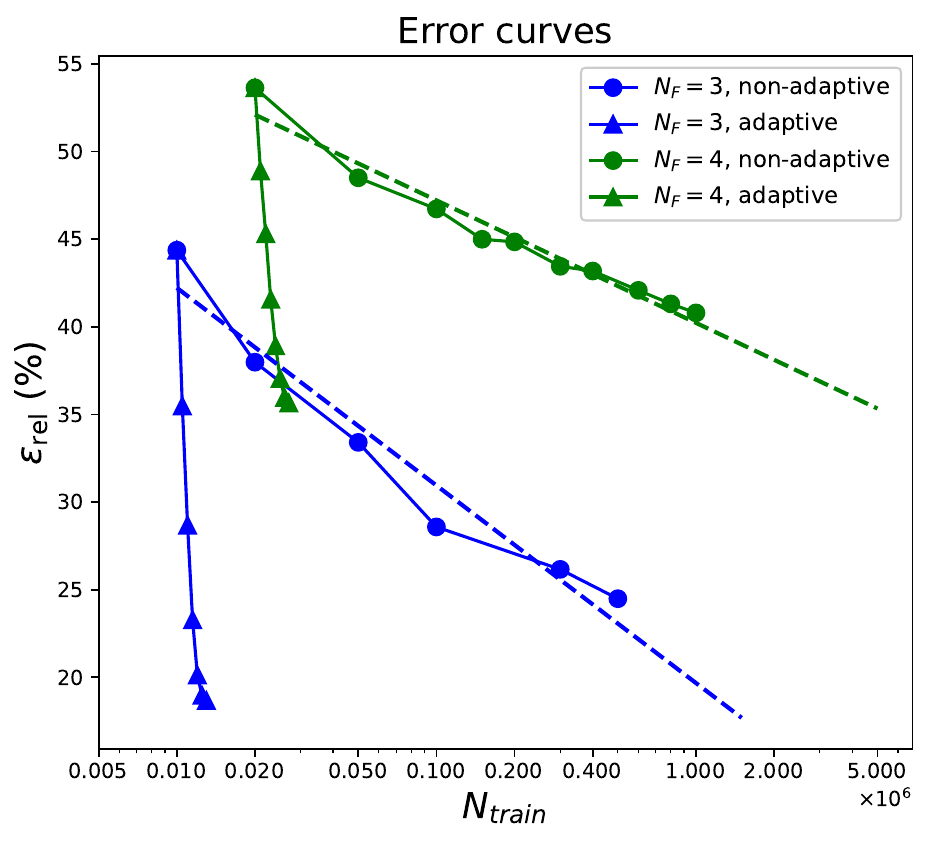}
\end{subfigure}
\hfill
\begin{subfigure}{0.49\textwidth}
    \centering
    \includegraphics[width=\textwidth]{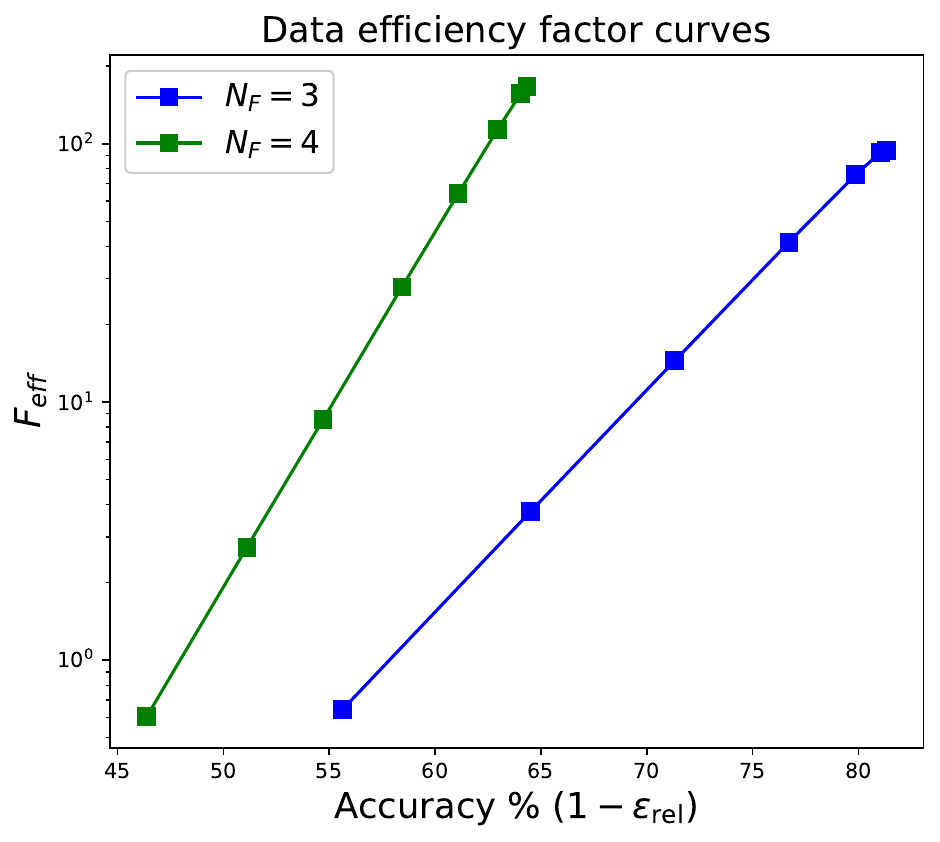}
\end{subfigure}
\end{subfigure}

 \caption{Data efficiency comparison between adaptive and non-adaptive training in the Fourier prior setting. \textbf{Left:} Average relative error versus training dataset size. \textbf{Right:} Data efficiency factor as a function of reached accuracy. See the caption of Figure~\ref{FIG:Circle curves} for further details.}
\label{FIG:Fourier curves}
\end{figure}

Similar to Figure~\ref{FIG:Circle curves}, Figure~\ref{FIG:Fourier curves} (left) compares the data scaling behavior of the standard non-adaptive one-shot training method and the adaptive sampling method. The fitted regression curve for the non-adaptive method again aligns closely with the actual data points, indicating that the scaling trend is well captured. The corresponding data efficiency factors, plotted against the target accuracy $1 - \varepsilon_\text{rel}$, are shown in the right panel of Figure~\ref{FIG:Fourier curves}.

As an example, consider the setting $N_F = 4$, with the adaptive method run for 7 rounds. For a single test instance, this results in a total of $N_{\text{base model}} + N_{\text{round}} \cdot N_{\text{adapt}} = 20000 + 7 \cdot 1000 = 27000$ training samples. At this cost, the adaptive method achieves a relative error of $35.6\%$, which matches the performance of a model trained on approximately $4494128$ samples in one shot. The resulting data efficiency factor is
\begin{equation*}
    F_{\text{eff}} = \frac{4494128}{27000} \approx 166.
\end{equation*}

Figure~\ref{FIG:Fourier curves} also includes results for the simpler case $N_F = 3$. The same pattern observed in the disk prior holds: as the prior becomes more complex, the non-adaptive method scales less favorably, while the adaptive method maintains much more consistent behavior. This leads to a steeper increase in data efficiency for more challenging prior.

Additionally, for the Fourier prior, we also compare the adaptive sampling approach with classical optimization-based methods; the results are shown in Table~\ref{TAB:Fourier GN}. The same trend observed under the disk prior persists: direct prediction from the adaptive model yields substantially lower reconstruction error and outperforms Gauss--Newton, even when initialized with the base-model prediction.

\begin{table}[!htb]
    \centering
    \begin{tabular}{l|cc}
    \toprule
        & $N_F=3$ & $N_F=4$ \\
    \midrule
        Direct prediction (base model) & 44.4\% & 53.6\%\\
        Direct prediction (adaptive model) & \textbf{18.7\%} & \textbf{35.7\%}\\
        Gauss--Newton (nearest-neighbor init.) & 97.6\% & 102.0\%\\
        Gauss--Newton (base-model init.) & 42.9\% & 47.8\%\\
    \bottomrule
    \end{tabular}
    \caption{Average reconstruction error on the test set under the Fourier prior, for direct prediction using the base and adaptive model (the latter trained on the adaptively constructed dataset), and for Gauss--Newton refinement under two initialization strategies. As in the disk prior case, direct prediction from the adaptive model outperforms both the base model and Gauss--Newton, even when the latter is initialized with the base-model prediction.}
    \label{TAB:Fourier GN} 
\end{table}

Finally, it is important to note that the reported number of training samples for the adaptive method corresponds to a single test instance. This holds for both the disk and Fourier prior settings, as well as for any other priors. When the method is applied to multiple test cases, the total cost of training data collection scales linearly with the number of instances, in contrast to the non-adaptive method, which trains a single model that is used for all test cases. However, in challenging regimes involving high-dimensional prior manifolds or high target accuracy, the non-adaptive method may require an enormous, or even unaffordable, amount of training data to produce a global model with barely acceptable accuracy. In such cases, the adaptive method remains effective by making progress on a per-instance basis and offers a practical advantage for solving complex inverse problems.

\section{Discussion and Future Work}

It should not be hard to see that for the adaptive sampling framework to work, we need the base model prediction ${\mathcal N}{\mathcal N}_{\theta_0}(\widehat m)$ to be within a useful range of the true inversion result ${\mathcal F}^{-1}(\widehat m)$. While this is inevitable, we observe in our numerical experiments that a suitably trained base model ${\mathcal N}{\mathcal N}_{\theta_0}$ usually does the work; see Section~\ref{SEC:Numerical}. Our adaptive framework is most efficient when training a performing base model ${\mathcal N}{\mathcal N}_{\theta_0}$ to start with is data- and cost-effective.

While the proposed adaptive sampling framework demonstrates strong performance in solving inverse scattering problems under structured priors, several important directions remain for future exploration. First, our current numerical experiments assume noiseless measurement data. Investigating robustness under various noise levels is, therefore, a natural next step. Second, the adaptive sampling strategy for learning the inverse map is not limited to the inverse scattering setup considered here; it can be readily applied to other inverse problems, such as wave inversion~\cite{wu2019inversionnet,DiReZh-HNA25}, or combined with classical approaches like the direct sampling method~\cite{ning2023direct,ning2025direct}. Finally, we have focused on priors defined by a manifold $\mathcal{M}$, relying on explicit structural assumptions. However, in many practical scenarios, prior information is more realistically described by a distribution or density supported on $\mathcal{M}$, which can be learned from data. Modeling such distributions based on available datasets allows us to go beyond rigid manifold assumptions, providing richer and more flexible prior information that may improve both sample efficiency and generalization. In this context, generative modeling techniques such as score-based diffusion models~\cite{sohl2015deep,ho2020denoising,song2021score} offer a promising direction for representing and sampling from complex prior or posterior distributions~\cite{chung2023diffusion,bruna2024provable,zhang2025back}. Exploring these extensions may further enhance the practicality and expressiveness of the adaptive sampling approach.


\section*{Acknowledgment}
We are grateful to Leslie Greengard and Manas Rachh for valuable discussions. The work of KR and NS is partially supported by the National Science Foundation through grants DMS-1937254 and DMS-2309802, and by the Gordon \& Betty Moore Foundation through award GBMF12801.

\vskip 0.2in
\bibliography{bib}
\bibliographystyle{siam}

\appendix
\section{Details of Two Types of Data Prior}

\subsection{Disk prior}
\label{APP:disk prior}
Below, we present a detailed description of the disk prior setting introduced in Section~\ref{SEC:disk prior}. In this setting, we assume prior knowledge that the true parameter $q^*$ belongs to a manifold $\mathcal{M}$ defined as follows. Positive integers $C_1 \leq C_2$, positive real numbers $r_1 \leq r_2$, and real numbers $A_1 \leq A_2$ specify the allowed ranges for the number, size, and amplitude of the disks.
\begin{enumerate}[(i)]
    \item An integer $C$ is drawn uniformly at random from $[C_1, C_2]$, representing the number of disks in $q^*$.

    \item For each of the $C$ disks, a radius and a center are uniformly sampled from $[r_1, r_2]$ and $\Omega$, respectively, ensuring the disks are disjoint and contained within $\Omega$ via rejection sampling.

    \item Each disk is assigned an amplitude randomly selected from $[A_1, A_2]$, resulting in a function $f$ over $\Omega$ formed by a linear combination of $C$ indicator functions of disjoint disks.
    
    \item The function $f$ is then smoothed by convolution with a Gaussian mollifier $\phi_\varepsilon$, yielding $q^* := f \ast \phi_\varepsilon$.
    
    \item Finally, $q^*$ is projected onto the space $\mathcal{Q}_{N}$.
\end{enumerate}
This defines the disk prior manifold $\mathcal{M}$. The projection onto $\mathcal{M}$ in line~\ref{alg:line_proj} and the local perturbation on $\mathcal{M}$ in line~\ref{alg:line_perturb} of Algorithm~\ref{ALGO:Adaptive Sampling} are provided in the main text.

\subsection{Fourier prior}
\label{APP:Fourier prior}
Below, we present a more detailed description of the Fourier prior setting introduced in Section~\ref{SEC:Fourier prior}.

In this setting, we assume prior knowledge that the true parameter $q^*$ belongs to a manifold $\mathcal{M}$ defined as follows. Recall that our domain is $\Omega = [-\pi/2, \pi/2]^2$. Fix a small $\varepsilon > 0$, and consider the smaller domain $\Omega' := [-\pi/2+ \varepsilon, \pi/2 - \varepsilon]$ with size $\pi-2\varepsilon$.
\begin{enumerate}[(i)]
    \item First, a random periodic function $f_1$ on $\Omega'$ is constructed using a truncated Fourier series with $N_F \times N_F$ modes. For $-N_F\le k,j \le N_F$, let $c_{k,j}$ and $d_{k,j}$ be independent samples from a standard normal distribution, and define
    \begin{equation*}
        f_1(x,y) := \Re\sum_{k=-N_F}^{N_F}\sum_{j=-N_F}^{N_F} (c_{k,j} + id_{k,j})\exp\left\{i\frac{2\pi}{\pi-2\varepsilon}(kx + jy)\right\}
    \end{equation*}
    for $(x,y)\in \Omega'$. Here, $i$ denotes the imaginary unit, and $\Re$ denotes taking the real part of the complex-valued expression.
    
    \item The function $f_1$ is rescaled to produce a physically meaningful wave speed profile by applying a piecewise linear map $\psi$, yielding $f_0 := \psi \circ f_1$. Specifically, $\psi$ maps $\min f_1$ to $\ell \sim \text{uniform}[-0.2, -0.1]$ (wave speed lower bound), $0$ to background wave speed, and $\max f_1$ to $h \sim \text{uniform}[2,3]$ (wave speed upper bound). The resulting function $f_0$ is then truncated to its first $N_F \times N_F$ Fourier modes.

    \item The truncated function $f_0$ is extended to the full domain $\Omega$ via $f := \chi_{\Omega'} f_0$ making sure it satisfies the zero boundary condition, and then smoothed by convolution with a Gaussian mollifier $\phi_\varepsilon$, yielding the final potential $q^* := f \ast \phi_\varepsilon$.

    \item Finally, $q^*$ is projected onto the space $\mathcal{Q}_{N}$.
\end{enumerate}
This defines the Fourier prior manifold $\mathcal{M}$.

In this setting, the projection onto $\mathcal{M}$ in line~\ref{alg:line_proj} and the local perturbation on $\mathcal{M}$ in line~\ref{alg:line_perturb} of Algorithm~\ref{ALGO:Adaptive Sampling} are implemented as follows. Given a parameter field $q$ on $\Omega$, consider its Fourier coefficients $c_{k,j}, d_{k,j}$ on the smaller domain $\Omega'$, so that 
\begin{equation*}
    q(x,y) = \sum_{k=-\infty}^{\infty}\sum_{j=-\infty}^{\infty} (c_{k,j} + id_{k,j})\exp\left\{i\frac{2\pi}{\pi-2\varepsilon}(kx + jy)\right\}
\end{equation*}
for $(x,y)\in \Omega'$. We denote these Fourier coefficients with the notation $\mathcal{F}_1 q(k,j) := c_{k,j}$ and $\mathcal{F}_2 q(k,j) := d_{k,j}$.

Given the prediction $\widehat{q}^{(t)}$ of the current neural network model $\mathcal{NN}$, we compute the Fourier coefficients $\mathcal{F}_1 \widehat{q}^{(t)}(k,j)$ and $\mathcal{F}_2 \widehat{q}^{(t)}(k,j)$ for $-N_F\le k,j\le N_F$. This corresponds to the projection onto the prior manifold in line 4 of Algorithm \ref{ALGO:Adaptive Sampling}. 

To sample around $\widehat{q}^{(t)}$ on $\mathcal{M}$, we perturb each Fourier coefficient by sampling from a normal distribution:
\begin{align*}
    \mathcal{F}_1^{\text{sample}}(k,j) &\sim \mathcal{N}(\mathcal{F}_1 \widehat{q}^{(t)}(k,j), [\sigma_1(k,j)]^2), \\
    \mathcal{F}_2^{\text{sample}}(k,j) &\sim \mathcal{N}(\mathcal{F}_2 \widehat{q}^{(t)}(k,j), [\sigma_2(k,j)]^2).
\end{align*}
We then reconstruct the perturbed field by repeating steps (i), (iii), and (iv) above (skipping the rescaling in step (ii)) using the new sampled Fourier coefficients.

The standard deviations $\sigma_1(k,j)$ and $\sigma_2(k,j)$ are estimated using the validation set predictions from the current model. Let $\{q_{\ell}^v\}_{{\ell}=1}^{N_v}$ denote the ground truth validation samples and $\{\widehat{q}_{\ell}^v\}_{\ell=1}^{N_v}$ their corresponding model predictions. Then,
\begin{align*}
    \sigma_1(k,j) &:= C_\sigma \cdot \frac{1}{N_v} \sum_{{\ell}=1}^{N_v} |\mathcal{F}_1 \widehat{q}_{\ell}^v(k,j) - \mathcal{F}_1 q_{\ell}^v(k,j)|, \\
    \sigma_2(k,j) &:= C_\sigma \cdot \frac{1}{N_v} \sum_{{\ell}=1}^{N_v} |\mathcal{F}_2 \widehat{q}_{\ell}^v(k,j) - \mathcal{F}_2 q_{\ell}^v(k,j)|.
\end{align*}
Here, the constant $C_\sigma > 1$ serves as a multiplicative factor to account for potential underestimation of uncertainty from the validation set, yielding a more conservative estimate of the perturbation scale. In our experiments, we set $C_\sigma = 2$.

\end{document}